\documentclass[10pt,twocolumn,letterpaper]{article}
\pdfoutput=1
\usepackage{cvpr}
\usepackage{times}
\usepackage{epsfig}
\usepackage{graphicx}
\usepackage{amsmath}
\usepackage{amssymb}

\usepackage{multirow}
\usepackage{array}
\newcolumntype{L}[1]{>{\raggedright\arraybackslash}m{#1}}
\newcolumntype{C}[1]{>{\centering\arraybackslash}p{#1}}
\usepackage{soul}
\usepackage{bbm}

\usepackage[pagebackref=true,breaklinks=true,letterpaper=true,colorlinks,bookmarks=false]{hyperref}

\cvprfinalcopy 

\def\cvprPaperID{6608} 

\begin{document}

\title{Exploit the Connectivity: Multi-Object Tracking with TrackletNet}

\author{Gaoang Wang, Yizhou Wang, Haotian Zhang, Renshu Gu, Jenq-Neng Hwang \vspace{0.5em}\\ 
University of Washington\\
{\tt\small \{gaoang, ywang26, haotiz, renshugu, hwang\}@uw.edu}
}

\maketitle

\begin{abstract}
   Multi-object tracking (MOT) is an important and practical task related to both surveillance systems and moving camera applications, such as autonomous driving and robotic vision. However, due to unreliable detection, occlusion and fast camera motion, tracked targets can be easily lost, which makes MOT very challenging. Most recent works treat tracking as a re-identification (Re-ID) task, but how to combine appearance and temporal features is still not well addressed. 
   In this paper, we propose an innovative and effective tracking method called 
   \textbf{TrackletNet Tracker (TNT)} that combines temporal and appearance information together as a unified framework. First, we define a graph model which treats each tracklet as a vertex. The tracklets are generated by appearance similarity with CNN features and intersection-over-union (IOU) 
   with epipolar constraints to compensate camera movement between adjacent frames. Then, for every pair of two tracklets, the similarity is measured by our designed multi-scale TrackletNet. Afterwards, the tracklets are clustered \cite{tang2018single} into groups which represent individual object IDs. 
   Our proposed TNT has the ability to handle most of the challenges in MOT, and achieve promising results on MOT16 and MOT17 benchmark datasets compared with other state-of-the-art methods.
\end{abstract}


\section{Introduction}


Multi-object tracking is an important topic in computer vision and machine learning field. This technique can be used in many tasks, such as traffic flow counting from surveillance cameras, human behavior prediction and autonomous driving assistance. 
However, due to noisy detections and occlusions, tracking multiple objects in a long time range is very challenging. To address such problems, many methods follow the tracking-by-detection framework, i.e., tracking is applied as an association approach given the detection results.
Built upon the tracking-by-detection framework, multiple cues can be combined together into the tracking scheme. 
1) Appearance feature of each detected object \cite{ristani2018features,zhang2017multi,tang2017multiple,wojke2017simple}. With a well-embedded appearance, features should be similar if they are from the same object, while they can be very different if they are from distinct objects.
2) Temporal relation for locations among frames in a trajectory \cite{milan2017online}. With slow motion and high frame rate of cameras, we can assume that the trajectories of objects 
are smooth in time domain. 
3) Interaction cue among different target objects which considers the relationship among neighboring targets 
\cite{sadeghian2017tracking}. 
As a result, we should take into account all these cues as an optimization problem. 

\begin{figure}[t]
\begin{center}
\includegraphics[width=\linewidth]{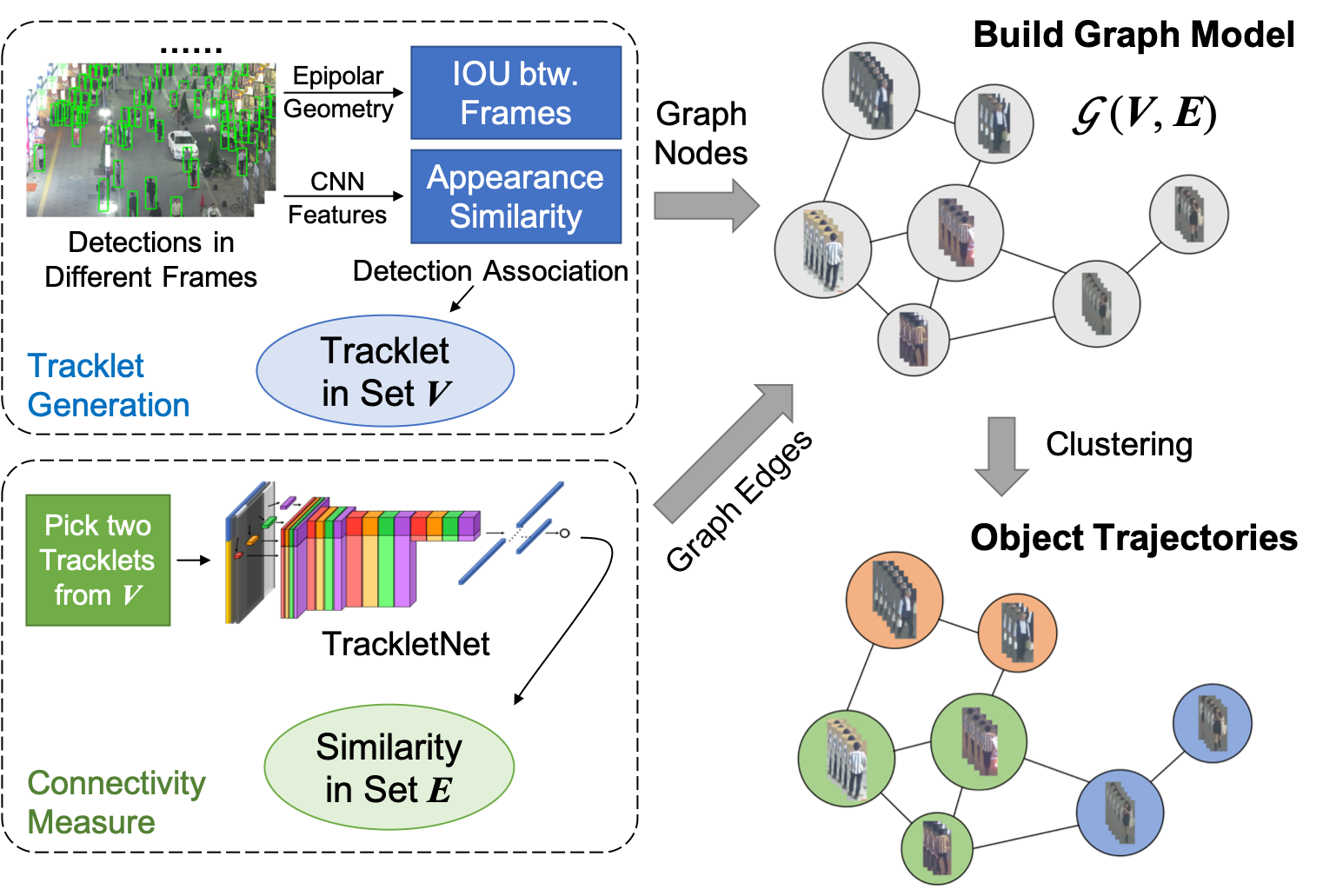}
\end{center}
   \caption{Our TNT framework for multi-object tracking. Given the detections in different frames, detection association is computed to generate Tracklets for the Vertex Set $V$. After that, each two tracklets are put into a novel TrackletNet to measure the connectivity, which formed the similarity on the Edge Set $E$. A graph model $\mathcal{G}$ can be derived from $V$ and $E$. Finally, the tracklets with the same ID are grouped into one cluster using the graph partition approach. 
   }
\label{fig:overall-flow}
\end{figure}

In this paper, the proposed TrackletNet Tracker (TNT) takes advantages of the above useful cues together into a unified framework based on an undirected graph model \cite{milan2016multi}. Each vertex in our graph model represents one tracklet and the edge between two vertices measures the connectivity
of two tracklets. Here, the tracklet is defined as a small piece of consecutive detections of an object. Due to the unreliable detections and occlusions, the entire trajectory of an object may be divided into several distinct tracklets.
Given the graph representation, tracking can be regarded as a clustering approach that groups the tracklets into one big cluster. 

To generate the tracklets, i.e., vertices of the graph, we associate detections among consecutive frames based on intersection-over-union (IOU) and the similarity of appearance features. However, the IOU criterion becomes unreliable because the position of detection may shift a lot when camera is moving or revolving. In such situation, epipolar geometry is adopted to compensate camera movement and predict the position of bounding boxes in the next frame. 
To estimate the connectivity on the edge of the graph between two vertices, the TrackletNet is designed for measuring the continuity of two input tracklets, which combines both trajectory and appearance information. 
The flowchart of our tracking method TNT is shown in Figure~\ref{fig:overall-flow}.

Specifically, we propose the following contributions:

1) We build a graph-based model that takes tracklets, instead of detected objects, as the vertices, to better utilize the temporal information and greatly reduce the computational complexity.

2) To the best of our knowledge, this is the first work to adopt epipolar geometry in tracklet generation to compensate camera movement. 

3) A CNN architecture, called multi-scale TrackletNet, is designed to measure the connectivity between two tracklets. This network combines trajectory and appearance information into a unified system. 

4) Our model outperforms state-of-the-art methods in multi-object tracking for both MOT16 and MOT17 benchmarks, and it can be also easily applied to other different scenarios.

\section{Related Work}

\paragraph{Graph Model based Tracking.} 
Most of the recent multi-object tracking approaches are based on tracking-by-detection schemes \cite{geiger20143d,zhang2013understanding}. Given detection results, we would like to associate detections across frames and estimate object locations when unreliable detection or occlusion occurs. 
Many tracking methods are based on graph models \cite{tang2017multiple,milan2016multi,tang2018single,tang2016multi,keuper2016multi,kumar2014multiple,choi2015near,tang2015subgraph,wen2014multiple} and solve the tracking problem by minimizing the total cost. 
In \cite{tang2017multiple,milan2016multi,tang2016multi,kumar2014multiple}, the detected objects are treated as the vertices in the graph models, while in \cite{tang2018single,choi2015near,tang2015subgraph,wen2014multiple}, the graph vertices are based on tracklets. 
For detection-based graph models, there are two major disadvantages. 
First, one of the important assumptions in graph models is the conditional independence of the vertices. However, detections are not conditional independent from frame to frame if we want to track an object in a long run. The temporal information is not well utilized.
Second, detection-based graph usually comes with a very high-dimensional affinity matrix, which makes it very hard to find the global minimum solution in the optimization. 
However, for tracklet-based graph models, it can better utilize the information from a short trajectory to measure the relationship between vertices, but the mis-association should be carefully handled in the tracklet generation step.

\paragraph{Tracking by RNN.} 
Besides graph models, recurrent neural networks (RNN)-based tracking also plays an important role in recent years \cite{sadeghian2017tracking,ma2018trajectory,kim2018multi,sadeghianend,lu2017online,milan2017online}. 
One advantage of RNN-based tracking is the ability of online prediction. 
However, along with the propagation of RNN block, the relation between two faraway detections becomes very weak. Without direct connections, the performance of RNN-based methods degrades in the long run and sometimes can be easily affected by unreliable detections.

\paragraph{Tracking by Feature Fusion.} 
Features are very important in the tracking-by-detection framework. There are two types of features that are used in common, i.e., appearance features and temporal features. 
For appearance features, many works adopt CNN-based features from Re-ID tasks \cite{ristani2018features,zhang2017multi,tang2017multiple}. However, histogram-based features, like color histograms, HOG, and LBP, are still powerful if no training data is provided \cite{tang2018single}. 
As for temporal features, the location, size, and motion of bounding boxes are commonly used. 
Given the appearance features and temporal features, the tracker can fuse them together using human defined weights \cite{zhang2017multi,milan2016multi,tang2018single}. Although \cite{sadeghian2017tracking,ma2018trajectory} propose RNN-based networks to combine features together, it is still empirical and difficult to determine the weight of each feature. 

\paragraph{End-to-End Tracking.} 
Another category of tracking is based on end-to-end frameworks \cite{feichtenhofer2017detect,kang2017t,kang2016object}, where we input raw video sequences and output object trajectory. In other words, the detection and tracking are trained jointly in a single-stage network. 
One major advantage of this framework is that the errors will not be accumulated from detection to tracking. The temporal information across frames can help improve the detection performance, while reliable detections can also feedback reliable tracking. However, such a framework requires a lot of training data. Without enough training data, overfitting becomes a severe problem. Unlike detection based training, tracking annotations for video sequences are usually hard to get, which becomes the major limitation of the end-to-end tracking framework.

\section{Tracklet Graph Model}
\label{sec:graph-model}





We use tracklets as the vertices in our graph model. Unlike the detection-based graph models, which are computational expensive and not well utilizing temporal information, we propose a tracklet-based graph model, which treats the tracklet as the vertex and measures the similarity between tracklets. From the tracklet, we can infer the object moving trajectory for a longer time, and we can also measure how the embedded features of the detections change along the time. Moreover, the number of tracklets is much less than the number of detections, which makes the optimization more efficiently.

In the following section, we will discuss in detail about the model parameters and optimization by tracklet clustering.

\subsection{Graph Definition $\mathbf{\mathcal{G}}(V,E)$}
\label{sec:graph-def}


\paragraph{Vertex Set.} 
A finite set $V$ in which every element $v \in V$ represents a tracklet of one object across multiple frames, i.e., a set of consecutive detections of the same object along time. For each detection, we define the bounding box with five parameters, i.e., the center 
of the bounding box $(x_t,y_t)$, the width and height $(w_t,h_t)$, and the frame index $t$. Besides the bounding box of the detection, we also extract an appearance feature \cite{schroff2015facenet} for each detected object at frame $t$. Note that because of unreliable detections, an entire trajectory of an object may be divided into multiple pieces of tracklets. The tracklet generation is explained in detail in Section~\ref{sec:epi-constri}.



\paragraph{Edge Set.} A finite set $E$ in which every element $e \in E$ represents an edge between two tracklets $u,w \in V$ that are not far away in the time domain, i.e., $\min_{t_u\in T(u), t_w \in T(w)} \lvert t_u - t_w \rvert \leq \delta_t$,
where $T(u)$ is the set of frame indices of the tracklet $u$. For tracklets that are far away, the edge is not considered between them since not enough information can be utilized for measuring their relationship. 

A connectivity measure $p_e$, represents the similarity of the two tracklets connected by the edge $e\in E$. The edge cost is defined as
\begin{equation}
    c = \log \left( \frac{1-p_e}{p_e} \right).
\label{eq:cost}
\end{equation}
Moreover, the connectivity is defined to be $0$ if two tracklets have overlap in the time domain since they must belong to distinct objects. This is because an object cannot appear in two tracklets at the same time. 
The connectivity is measured by our designed TrackletNet, which will be introduced in Section~\ref{sec:trackletnet}.

\subsection{Tracklet Clustering}


After the tracklet graph is built, we acquire the object trajectories by clustering the graph into different sub-graphs. The tracklets in each sub-graph can represent the same object. We will explain some details of our tracklet clustering in the following paragraphs.

\paragraph{Feasible Solutions.} 
Given a tracklet graph $\mathcal{G}(V,E)$, we hope to partition $\mathcal{G}$ into disjoint sub-graphs $\mathcal{G}[s_{\tau}]$, and each sub-graph represents a distinct object. Here $\forall s_{\tau} \subseteq V$, $\tau$ represents the object ID.
Thus, every tracklet $u \in s_{\tau}$ is from the same object $\tau$ and any two tracklets $u \in s_{\tau}, w \in s_{\tau'}$ from two different sub-graphs are from different objects $\tau$ and $\tau'$. For the graph partition problem, the global optimal solution cannot be easily guaranteed. But we can still define the feasible solutions as follows.
\begin{itemize}
\setlength\itemsep{-0.2em}
    \item Each sub-graph $\mathcal{G}[s_{\tau}]$ should be a connected graph, i.e., $\forall \tau$,  $\forall u,w \in s_{\tau}$, $\exists P\in \mathcal{G}[s_{\tau}]$, s.t., $u,w\in P$, where $P$ is a path inside $\mathcal{G}[s_{\tau}]$.
    \item The cost on the edge inside each sub-graph should have a finite value, i.e., $\forall \tau$, $\forall u,w \in s_{\tau}$, if $\exists e\in E$ for $u,w$, $p_{e}\neq0$.
\end{itemize}

\paragraph{Objective Function.} 
The objective function is defined to minimize the total clustering cost on all graph edges. We define $\pi(u,w)\in\{\pm 1\}$ as the clustering label for tracklets $u$ and $w$. If $u$ and $w$ are partitioned into one sub-graph, $\pi(u,w)$ is set to be $+1$; otherwise, $\pi(u,w)$ is set to be $-1$. The objective function is defined as follows,
\begin{equation}
    \mathcal{O} = \min_{\pi \in \{\pm 1\}}\sum_{\substack{u,w \in V\\u \in N(w)}} \pi(u,w) \cdot c(u,w),
\label{eq:cluster-opt}
\end{equation}
where $N(w)$ represents the set of neighboring tracklets of $w$ with edge shared in the graph.

\paragraph{Clustering.} 
The graph partition is formulated as a clustering problem. However, the minimum cost of graph cut problem defined by Equation~(\ref{eq:cluster-opt}) is APX-hard \cite{papadimitriou1991optimization}. 
Besides, the number of clusters is unknown in advance. 
In this work, we exploit a greedy search-based clustering method proposed by \cite{tang2018single} to minimize the cost. Five clustering operations, i.e., assign, merge, split, switch, and break, are used. The advantage of adopting different types of clustering operation is to avoid being stuck at the local minimum as much as possible in the optimization.

\section{Proposed TrackletNet Tracker} 

\subsection{Tracklet Generation with Epipolar Constraints}
\label{sec:epi-constri}

As defined in Section~\ref{sec:graph-model}, a tracklet contains consecutively detected objects with bounding box information and appearance features with dimension $d_{ap}$. To simplify the generation of tracklets, we associate two consecutive detections based on IOU and appearance similarity in adjacent frames with a high association threshold to guarantee the mis-association as small as possible \cite{zhang2017multi,wang2016closed}. 



\begin{figure}
\begin{center}
\includegraphics[width=\linewidth]{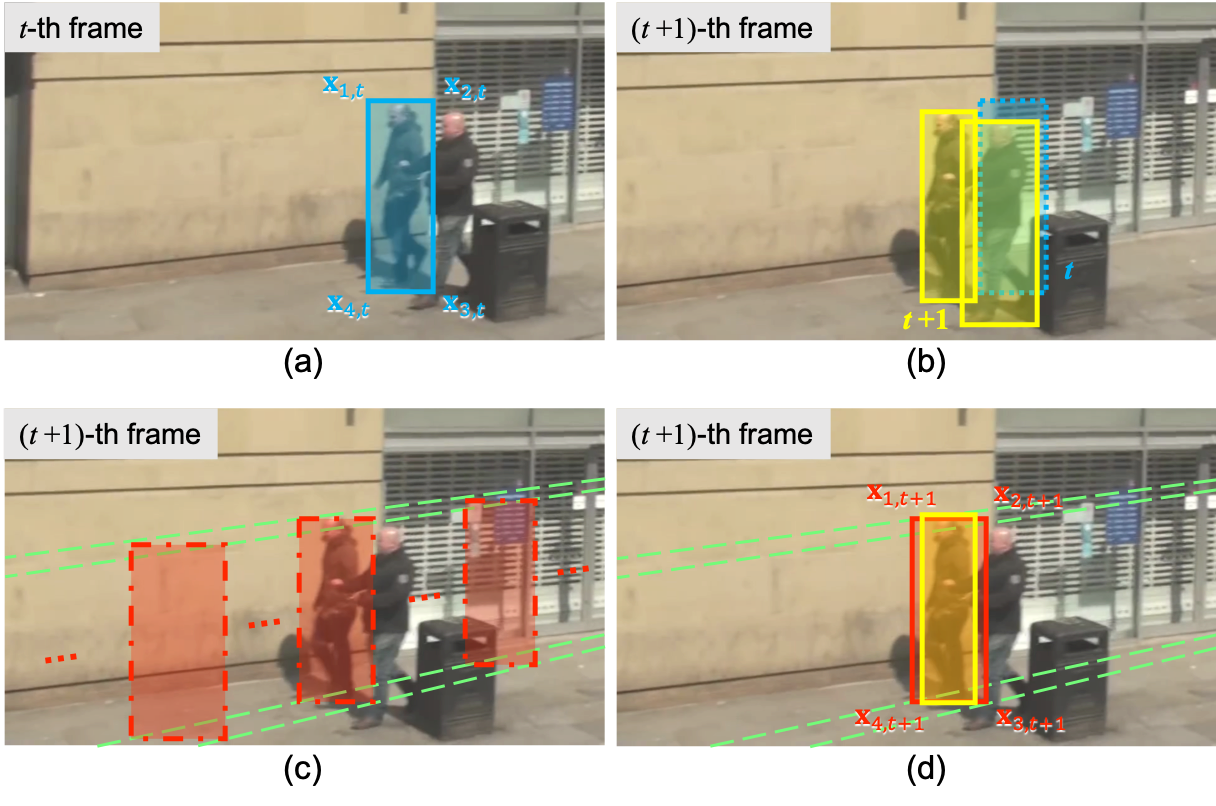}
\end{center}
   \caption{An example of EG-based detection association. 
   (a) $t$-th frame with target detection (blue). 
   (b) ($t$+1)-th frame with new detections (yellow).
   The target detection from $t$-th frame (blue dash-box) has a larger IOU with a different candidate detection in ($t$+1)-th frame (right yellow box). 
   (c) examples of candidate predicted bounding boxes (red dash-boxes) intersected with epipolar lines (green dash-lines). 
   (d) the predicted bounding box (red) in ($t$+1)-th frame overlapped with the correct detection (yellow). }
\label{fig:epi-geo}
\end{figure}

However, the association accuracy can still be affected by the fast motion of the camera. For example, as shown in the Figure~\ref{fig:epi-geo}(a)(b), the target detection in the $t$-th frame has a large IOU with another detection in the $(t+1)$-th frame. As a result, the detection may easily get mis-associated.

This issue can be well solved by epipolar geometry (EG) \cite{hartley2003multiple}, i.e., $\mathbf{x}_{t}^\top \mathbf{F} \mathbf{x}_{t+1}=0$ for any matched static feature point $\mathbf{x}$ in two frames, where $\mathbf{F}$ is the fundamental matrix.
First, if we assume the target is static or has slow motion, then the four corner points $\mathbf{x}_{i,t}$ of the target detection bounding box in the $t$-th frame should lie on the corresponding epipolar lines in the $(t+1)$-th frame, i.e., the predicted target bounding box in the $(t+1)$-th frame should intersect with the four epipolar lines as much as possible as shown in Figure~\ref{fig:epi-geo}(c). Second, we also assume the size of the bounding box does not have much change in adjacent frames, then the optimal predicted bounding box can be obtained, which is shown in red in Figure~\ref{fig:epi-geo}(d). 

Followed by the above two assumptions, we can predict the target bounding box location in the $(t+1)$-th frame by formulating an optimization problem. 
Define four corner points of the target bounding box in the $t$-th frame as $\mathbf{x}_{i,t}$, where $i\in\{1,2,3,4\}$, like the example shown in Figure~\ref{fig:epi-geo}(a). 
Similarly, we define $\mathbf{x}_{i,t+1}, i \in \{1,2,3,4\}$, as the bounding box in the $(t+1)$-th frame. 
Then we can define the cost function as follows,
\begin{equation}
\label{eq:epi-geo-cost}
\begin{aligned}
    f(\mathbf{x}_{i,t+1}) = &\sum_{i=1}^4 \lVert \mathbf{x}_{i,t}^\top \mathbf{F} \mathbf{x}_{i,t+1} \rVert_2^2 \\
    + &\lVert (\mathbf{x}_{3, t+1} - \mathbf{x}_{1,t+1}) - (\mathbf{x}_{3,t} - \mathbf{x}_{1,t} ) \rVert_2^2,
\end{aligned}
\end{equation}
where the first term guarantees the predicted bounding box should intersect with four corresponding epipolar lines as much as possible, while the second term is the target size constraint. One example of predicted bounding box, as shown in Figure~\ref{fig:epi-geo}(d), is well aligned with the true target in $(t+1)$-th frame. Then, in the detection association, IOU is calculated between predicted bounding boxes 
and detection bounding boxes in the $(t+1)$-th frame.


Fundamental matrix $\mathbf{F}$ can be estimated by the RANSAC \cite{fischler1981random} algorithm with matched SURF points \cite{bay2006surf} between two consecutive frames. 

The optimization of the cost function in Equation~(\ref{eq:epi-geo-cost}) can be reformulated into a Least Square problem and solved efficiently.


\subsection{Multi-Scale TrackletNet}
\label{sec:trackletnet}

\begin{figure*}
\begin{center}
\includegraphics[width=0.95\linewidth]{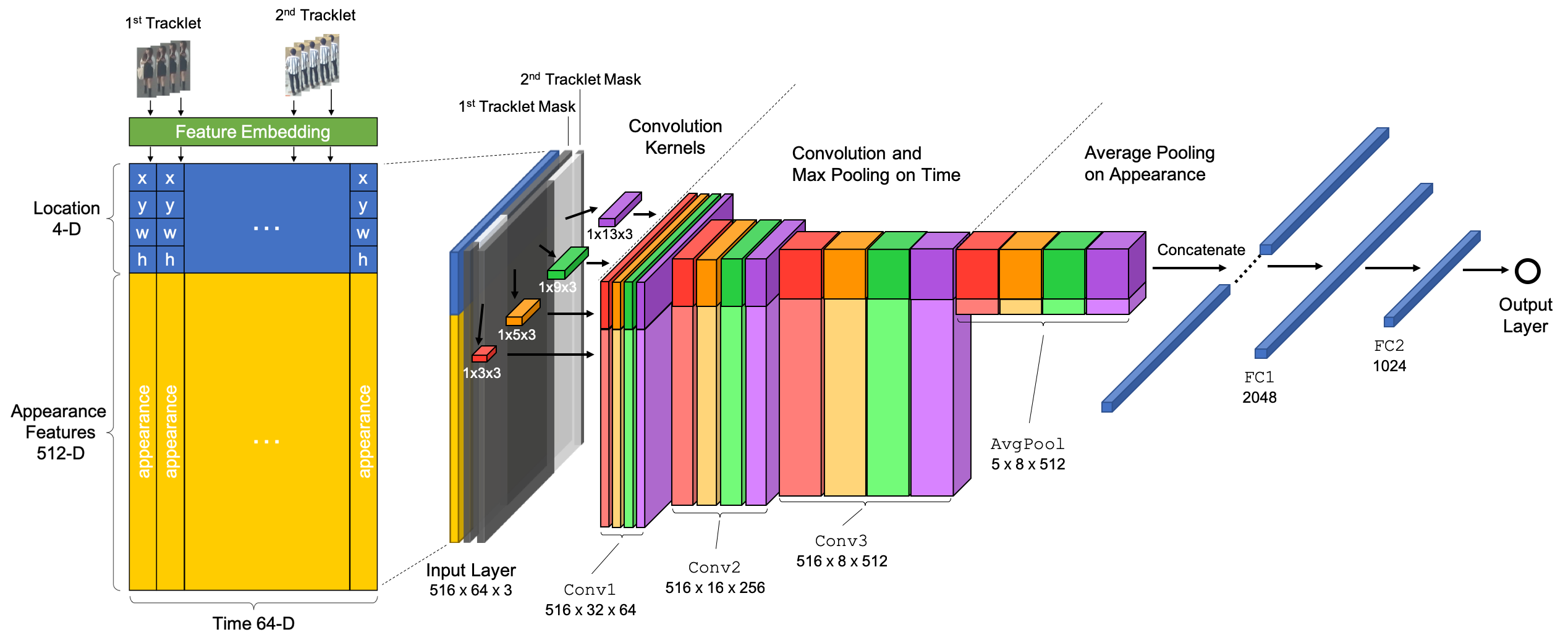}
\end{center}
   \caption{Architecture of Multi-scale TrackletNet. 
   First, we extract embedded features from two input tracklets, which include $4D$ location features and $512D$ appearance features along the time window of $64$ frames.
   The input tensor has three channels, i.e., one for tracklet embedded features and the other two for binary masks, where white color represents 1 and black color represents 0. Four types of $1D$ convolution kernels are applied for feature extraction in three convolution layers. 
   For each convolution layer, max pooling is adopted for down-sampling in the time domain. Average pooling is conducted on the dimensions of the appearance feature after \texttt{Conv3}. Then two fully connected layers are conducted to get the final output. 
   }
\label{fig:trackletnet}
\end{figure*}


To measure the connectivity between two tracklets,
we aggregate different types of information, including temporal and appearance features via the designed multi-scale TrackletNet. 
The architecture of the proposed TrackletNet is shown in Figure~\ref{fig:trackletnet}.

For each frame $t$, a vector consisting of the bounding box parameters, i.e., $(x_t$, $y_t$, $w_t$, $h_t)$, concatenated by an embedded appearance feature extracted from the FaceNet \cite{schroff2015facenet}, is used to represent an individual detection from a tracklet. Considering two tracklets with edge-shared in the graph, we concatenate the embedded feature of each detection from these two tracklets inside a time window with a fixed size $T$. Then the feature space in the time window of the two tracklets is $(4+d_{ap}) \times T$. As for frames between the two target tracklets, we use a $(4+d_{ap})$ dimensional interpolated vector instead at each missing frame $t$. Besides, zero-padding is used for frames after the second tracklet. 
To better represent the time duration of input tracklets, two binary masks are used as individual channels with $(4+d_{ap}) \times T$ dimension for each input tracklet. For each frame $t$, if the detection exists, then we set the $t$-th column of the binary mask to be all $\mathbf{1}$ vector; otherwise we set $\mathbf{0}$ vector instead. 
As a result, the size of the input tensor of the TrackletNet is $B \times (4+d_{ap}) \times T \times 3$, where $B$ is the batch size and $3$ indicates the number of channels, one for the embedded feature space and the other two for the binary masks. 

TrackletNet contains three convolution layers \texttt{Conv1}, \texttt{Conv2}, \texttt{Conv3}, one average pooling layer \texttt{AvgPool}, and two fully connected layers \texttt{FC1}, \texttt{FC2}. 
For each convolution layer, four different sizes of kernels are used, i.e., $1 \times3$, $1 \times 5$, $1 \times 9$, $1 \times 13$. Note that our convolution is only in the time domain, which can measure the continuity for each dimension of the feature. Different sizes of kernels will look for feature changes in different scales. The large kernels have the ability to measure the continuity of two tracklets even if they are far away in the time domain, while small kernels can focus on appearance difference if input tracklets are in small pieces. Each convolution is followed by one max pooling layer which down-samples by $2$ in the time domain. After \texttt{Conv3}, we take the average pooling on appearance feature dimensions. \texttt{AvgPool} plays a role of the weighted majority vote on the discontinuity of all appearance dimensions. Then we concatenate all features and use two fully connected layers for the final output. The output is defined as the similarity between the two input tracklets, which ranges from zero to one. 



There are some important properties of the TrackletNet, which are listed as follows.
\begin{itemize}
\setlength\itemsep{-0.2em}
    \item TrackletNet focuses on the continuity of the embedded features along the time. Because of the independence among different feature dimensions, no convolution is conducted across the dimensions of the embedded features. In other words, the convolution kernels only capture the dependency along time.
    \item Binary masks of the input tensor play a role as the tracklet indicator, telling the temporal locations of the tracklets. They help the network learn if the discontinuity of two tracklets is caused by frames without detection or the abrupt changes of the tracklets. 
    \item The network integrates object Re-ID, temporal and spatial dependency as one unified framework.
\end{itemize}


\section{Experiments}
\label{sec:experiment}

\subsection{Dataset}
\label{sec:dataset}
We use MOT16 and MOT17 \cite{milan2016mot16} datasets to train and evaluate our tracking performance. For MOT16 dataset, there are $7$ training video sequences and $7$ testing video sequences. The benchmark also provides public deformable part models (DPM) \cite{felzenszwalb2010object} detections for both training and testing data. MOT17 has the same video sequences 
as MOT16 but provides more accurate ground truth in the evaluation. In addition to DPM, Faster-RCNN \cite{ren2015faster} and scale dependent pooling (SDP) \cite{yang2016exploit} detections are also provided for evaluating the tracking performance. The number of trajectories in the training data is $546$ and the number of total frames is $5,316$.

\subsection{Implementation Details}


Our proposed multi-scale TrackletNet is purely trained on MOT dataset. The extracted appearance feature has $512$ dimensions, i.e., $d_{ap}=512$. The time window $T$ is set to $64$ and the batch size $B$ is set to $32$. We use Adam optimizer with a learning rate of $10^{-3}$ at the beginning. We decrease the learning rate by $10$ times for every $2,000$ steps until it reaches $10^{-5}$. As mentioned above, the MOT dataset is quite small for training a complex neural network. However, the framework of our proposed TNT is carefully designed to avoid over-fitting. In addition, augmentation approaches are used for generating the training data, i.e., tracklets, as follows. 


\paragraph{Bounding box randomization.} Instead of using the ground truth bounding boxes for training, we randomly disturb the size and location of bounding boxes 
by a factor $\alpha$ sampled from the normal distribution $\mathcal{N}(0,0.05^2)$. Since the detection results could be very noisy, this randomization will make sure the data from training and testing are as similar as possible. For each embedded detection before TrackletNet, the four parameters, i.e., $(x,y,w,h)$, are normalized by the size of the frame image to ensure the input of TrackletNet keeps the same scale in different datasets.

\paragraph{Tracklet generation.} Here, we randomly divide the trajectory of each object into small pieces of tracklets as follows. For each frame, we sample a random number from the uniform distribution, if it is smaller than a threshold, then we set this frame as the breaking frame. Then we split the entire trajectory based on the breaking frames into tracklets.

In the training stage, we randomly generate tracklets with augmentations mentioned above. For each training data, two tracklets are randomly selected as the input if they can satisfy the condition of the edge defined in the graph model in Section~\ref{sec:graph-def}. If they are from the same object, the training label is set to be $1$; otherwise, $0$ is assigned as the label. To make it no bias, positive and negative pairs are sampled equally.



\subsection{Feature Map Visualization}

\begin{figure}
\begin{center}
\includegraphics[width=\linewidth]{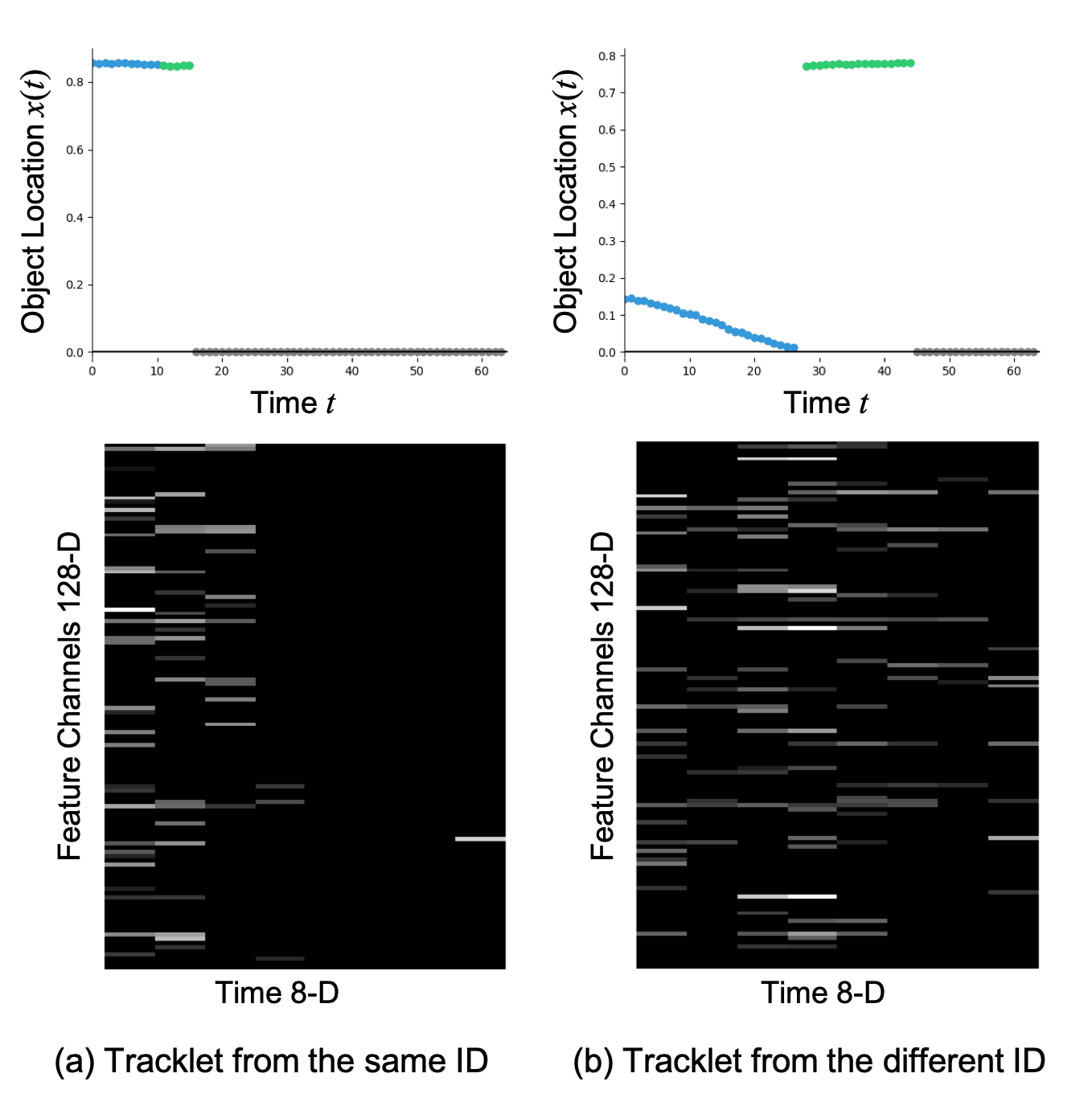}
\end{center}
   \caption{Examples of feature maps. For each column, the top figure shows the spatial locations of the two input tracklets in the $64$-frame time window. The bottom figure is the corresponding feature map after the max pooling of \texttt{Conv3} with the kernel size $5$, which aligns with the figure in the top row in the time domain. We can see that the connection part of the input tracklets in the time domain have strong activations.
   }
\label{fig:feature-map} 
\end{figure}

To better understand the effectiveness of our proposed TrackletNet, we also plot two examples of feature maps as shown in Figure~\ref{fig:feature-map}. For each column (a) and (b) in Figure~\ref{fig:feature-map}, the top figure shows the spatial locations of the two input tracklets in the $64$-frame time window. Blue and green colors represent two tracklets respectively. The bottom figure shows the corresponding feature map in the time-channel plane after the max pooling of \texttt{Conv3} with kernel size $5$. The horizontal axis represents the time domain which aligns with the figures in the top row, while the vertical axis represents different channels in the feature map. For the example shown in (a), most higher values of the feature map are on the left side since the connection between the two tracklets is on the left part of the time window. As for (b), higher values in the feature map are on the middle side of the time window, which also matches the situation of the two input tracklets. From the feature map, we can see that the connection part of the input tracklets has strong activation, which is critical for the connectivity measurement. 

\subsection{Tracking Performance}

\paragraph{Quantitative results on MOT16 and MOT17 datasets.}

We also provide our quantitative results on MOT16 and MOT17 benchmark datasets compared with other state-of-the-art methods, which are shown in Table~\ref{tab:mot16} and Table~\ref{tab:mot17}. Note that we use IDF1 \cite{ristani2016performance} and MOTA as major factors to evaluate the reliability of a tracker. As mentioned in \cite{ristani2016performance}, there are several weaknesses of MOTA metric, which is very sensitive to the detection threshold. Instead, IDF1 score compares ground truth trajectory and computes trajectory by a bipartite graph, which reflects how long of an object has been correctly tracked. 
We can see that our IDF1 score is much higher than other state-of-the-art methods. For other metrics shown in the table, we are also among the top rankings. 

\begin{figure}
\begin{center}
\includegraphics[width=\linewidth]{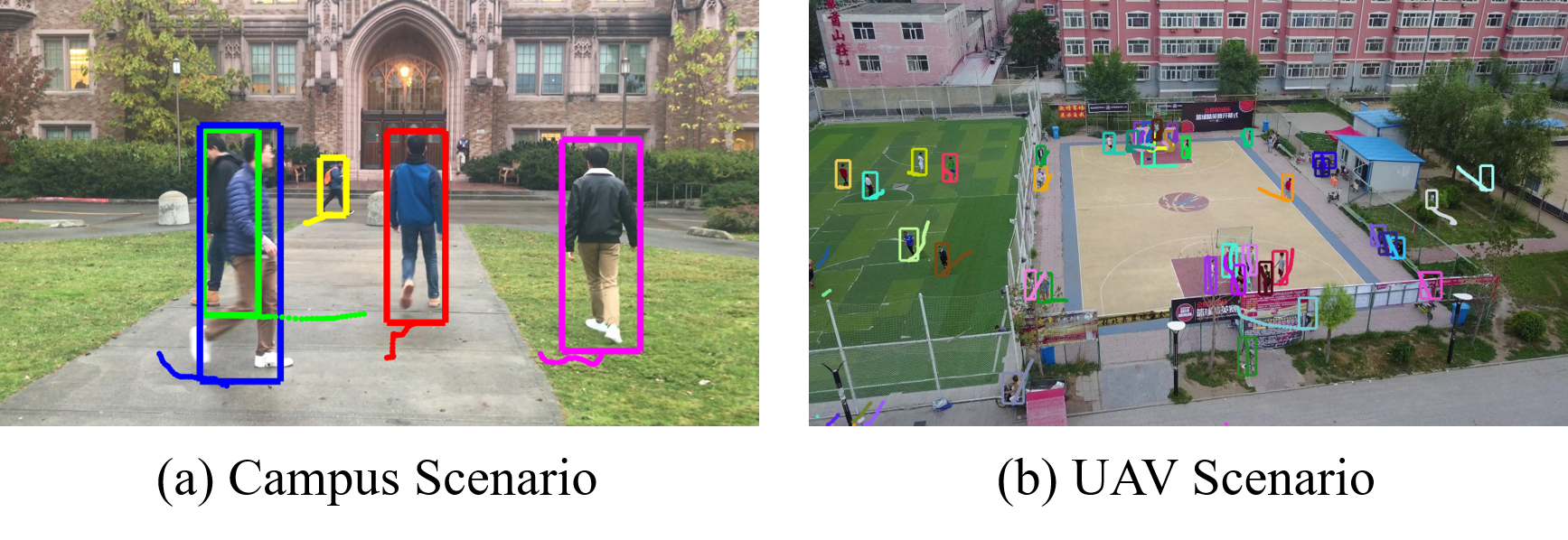}
\end{center}
   \caption{Tracking in different scenarios. (a) Tracking on campus pose estimation dataset. 3D human pose can be further estimated using the tracking results. (b) Tracking for UAV applications.
   }
\label{fig:app_example} 
\end{figure}

\paragraph{Qualitative results for different scenarios.}
With the trained model on the MOT dataset, we also test our proposed tracker on other scenarios without any fine-tuning. Promising results are also achieved. Figure~\ref{fig:app_example} shows some qualitative tracking results using our proposed tracker on other applications, like 3D pose estimation and UAV applications. 

\begin{table*}
\begin{center}
\begin{tabular}{|L{2.9cm}|C{1.2cm}|C{1.3cm}|C{1.2cm}|C{1.2cm}|C{1.2cm}|C{1.2cm}|C{1.2cm}|C{1.2cm}|}
\hline
Tracker & IDF1 $\uparrow$ & MOTA $\uparrow$ & MT $\uparrow$ & ML $\downarrow$ & FP $\downarrow$ & FN $\downarrow$ & IDsw. $\downarrow$ & Frag $\downarrow$ \\
\hline\hline
GCRA \cite{ma2018trajectory} & 48.6 & 48.2 & 12.9\% & 41.1\% & \textbf{5,104} & 88,586 & 821 & 1,117 \\
oICF \cite{kieritz2016online} & 49.3 & 43.2 & 11.3\% & 48.5\% & \noindent\color{cyan}{6,651} & 96,515 & 381 & 1,404 \\
MOTDT \cite{long2018real} & 50.9 & 47.6 & 15.2\% & \textbf{38.3\%} & 9,253 & \noindent\color{cyan}{85,431} & 792 & 1,858 \\
LMP \cite{tang2017multiple} & 51.3 & \noindent\color{cyan}{48.8} & 18.2\% & \noindent\color{cyan}{40.1\%} & 6,654 & 86,245 & 481 & \noindent\color{cyan}{595} \\
MCjoint \cite{keuper2016multi} & 52.3 & 47.1 & \textbf{20.4\%} & 46.9\% & 6,703 & 89,368 & \noindent\color{cyan}{370} & 598 \\
NOMT \cite{choi2015near} & 53.3 & 46.4 & \noindent\color{cyan}{18.3\%} & 41.4\% & 9,753 & 87,565 & \textbf{359} & \textbf{504} \\
DMMOT \cite{zhu2018online} & \noindent\color{cyan}{54.8} & 46.1 & 17.4\% & 42.7\% & 7,909 & 89,874 & 532 & 1,616 \\
\hline
\textbf{TNT} (Ours) & \textbf{56.1} & \textbf{49.2} & 17.3\% & 40.3\% & 8,400 & \textbf{83,702} & 606 & 882 \\
\hline
\end{tabular}
\end{center}
\caption{Tracking performance on the MOT16 test set. Best in bold, second best in blue. 
}
\label{tab:mot16}
\end{table*}

\begin{table*}
\begin{center}
\begin{tabular}{|L{2.9cm}|C{1.2cm}|C{1.3cm}|C{1.2cm}|C{1.2cm}|C{1.2cm}|C{1.2cm}|C{1.2cm}|C{1.2cm}|}
\hline
Tracker & IDF1 $\uparrow$ & MOTA $\uparrow$ & MT $\uparrow$ & ML $\downarrow$ & FP $\downarrow$ & FN $\downarrow$ & IDsw. $\downarrow$ & Frag $\downarrow$ \\
\hline\hline
MHT\_DAM \cite{kim2015multiple} & 47.2 & 50.7 & 20.8\% & 36.9\% & \noindent\color{cyan}{22,875} & 252,889 & 2,314 & \textbf{2,865} \\
FWT \cite{henschel2018fusion} & 47.6 & \noindent\color{cyan}{51.3} & 21.4\% & \textbf{35.2\%} & 24,101 & 247,921 & 2,648 & 4,279 \\
HAM\_SADF17 \cite{yoon2018online} & 51.1 & 48.3 & 17.1\% & 41.7\% & \textbf{20,967} & 269,038 & \noindent\color{cyan}{1,871} & 3,020 \\
EDMT17 \cite{chen2017enhancing} & 51.3 & 50.0 & \noindent\color{cyan}{21.6\%} & 36.3\% & 32,279 & \noindent\color{cyan}{247,297} & 2,264 & 3,260 \\
MOTDT17 \cite{long2018real} & 52.7 & 50.9 & 17.5\% & 35.7\% & 24,069 & 250,768 & 2,474 & 5,317 \\
jCC \cite{keuper2018motion} & 54.5 & 51.2 & 20.9\% & 37.0\% & 25,937 & 247,822 & \textbf{1,802} & 2,984 \\
DMAN \cite{zhu2018online} & \noindent\color{cyan}{55.7} & 48.2 & 19.3\% & 38.3\% & 26,218 & 263,608 & 2,194 & 5,378 \\
\hline
\textbf{TNT} (Ours) & \textbf{58.0} & \textbf{51.9} & \textbf{23.5\%} & \noindent\color{cyan}{35.5\%} & 37,311 & \textbf{231,658} & 2,294 & \noindent\color{cyan}{2,917} \\
\hline
\end{tabular}
\end{center}
\caption{Tracking performance on the MOT17 test set. Best in bold, second best in blue. 
}
\label{tab:mot17}
\end{table*}

\subsection{Ablation Study} 

\paragraph{Occlusion Handling.}

\begin{figure}
\begin{center}
\includegraphics[width=\linewidth]{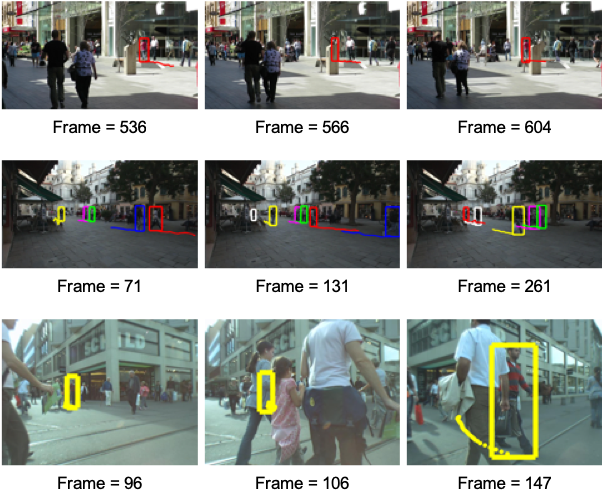}
\end{center}
    \caption{Occlusion handling in different MOT sequences. }
\label{fig:occlusion}
\end{figure}

Occlusion is one of the major challenges in MOT. Our framework can easily handle both partial and full occlusions even with a very long time range. 
When a person is occluded, the detection as well as appearance features are unreliable. 
During generating the tracklets, when we test that there is a large change in appearance, we just stop detection association even the detection result is available. After several 
or tens of frames, when the same person appears again from occlusion, a new tracklet will be assigned to the person. Then the connectivity between these two tracklets will be measured to distinguish whether they are the same person. Once they are confirmed with the same ID, we can easily fill out the missing detections with linear interpolation. Figure~\ref{fig:occlusion} shows qualitative results for handling occlusions. The first row of Figure~\ref{fig:occlusion} is from the MOT17-08 sequence. At frame $566$, the person with a red bounding box is fully occluded by a statue. But it can be correctly tracked after it appears again at frame $604$. The second row is one example of the MOT17-01 sequence, the person with the red bounding box goes across five other pedestrians, but the IDs of all targets keep consistent along the time. The last row shows the person with a yellow bounding box is crossing the street from MOT17-06 sequence captured with a moving camera. Although it is occluded by several other pedestrians, it can be still effectively tracked in a long run.

\paragraph{Effectiveness of Tracklet Generation with Epipolar Geometry.}

To check the effectiveness of EG in tracklet generation, we run detection association on MOT17-10 and MOT17-13 with the Faster-RCNN detector because these two sequences have large camera motion. Table~\ref{tab:epipolar-geometry} shows the results with/without epipolar geometry. Two types of error rates are evaluated, i.e., false discovery rate (FDR) and false negative rate (FNR), which are defined as follows, 
\begin{equation}
    \text{FDR} = \frac{\text{FP}}{\text{TP} + \text{FP}},\ \ 
    \text{FNR} = \frac{\text{FN}}{\text{TP} + \text{FN}},
\end{equation}
where TP, FP and FN represent true positive, false positive and false negative, respectively. 

From Table~\ref{tab:epipolar-geometry}, we can see that FDR is quite small in both cases, which means only a small portion of incorrect associations is involved in the tracklet generation. It shows the effectiveness of our tracklet-based graph model. On the other hand, FNR largely drops with epipolar geometry adopted, especially for the MOT17-13 sequence, which reflects the effectiveness of the proposed tracklet generation strategy.

\begin{table}
\begin{center}
\begin{tabular}{|l|c|c|c|}
\hline
Video Seq. & EG Involved & FDR (\%) & FNR (\%) \\
\hline\hline
\multirow{2}*{MOT17-10} & $\times$ & 2.4 & 6.5 \\
～ & \checkmark & \textbf{2.4} & \textbf{5.9} \\
\hline
\multirow{2}*{MOT17-13} & $\times$ & 3.6 & 12.4 \\
~ & \checkmark & \textbf{3.4} & \textbf{9.7} \\
\hline
\end{tabular}
\end{center}
\caption{The effectiveness of tracklet generation with EG.} 
\label{tab:epipolar-geometry}
\end{table}

\paragraph{Robustness to Appearance Features.}

Another major advantage of our TrackletNet is the ability to address overfitting learning of appearance features. Different from \cite{ma2018trajectory}, our TrackletNet is trained only on MOT dataset without using additional tracking datasets, but we can still achieve very good performance. This is because of the dimension independence of appearance features in training the network with convolutions only conducted in the time domain. As a result, the complexity of the network is largely reduced, which also decreases the effect of overfitting. 

\begin{table}
\begin{center}
\begin{tabular}{|c|C{1.8cm}|c|c|c|}
\hline
Noise (Std) & Method & IDF1 & MOTA & IDsw. \\
\hline\hline
\multirow{2}*{$\sigma=$ 0.05} & Baseline & 31.7 & 22.4 & 23 \\
～ & \textbf{TNT} & \textbf{34.1} & \textbf{22.5} & \textbf{20} \\
\hline
\multirow{2}*{$\sigma=$ 0.1} & Baseline & 31.1 & 22.1 & 26 \\
~ & \textbf{TNT} & \textbf{34.1} & \textbf{22.3} & \textbf{21} \\
\hline
\multirow{2}*{$\sigma=$ 0.2} & Baseline & 20.6 & 19.0 & 80 \\
~ & \textbf{TNT} & \textbf{34.0} & \textbf{22.5} & \textbf{20} \\
\hline
\end{tabular}
\end{center}
\caption{The robustness of TNT compared with the baseline method to disturbed appearance features. 
}
\label{tab:robust}
\end{table}

To test the model robustness to appearance features, we disturb the appearance features with Gaussian noise on MOT17-02 sequence. The compared baseline method is using the Bhattacharyya distance of appearance features between the input pair of tracklets as the edge cost in the graph, which is commonly used in person Re-ID tasks. 
The comparison results are shown in Table~\ref{tab:robust} with Gaussian noise using different standard deviations (Std). From the table, we can see that the baseline method degrades largely with the increasing of noise level, while the tracking performance is not affected much for TNT. This is because TNT measures the temporal continuity of features as the similarity rather than using feature distance itself, which can largely suppress unreliable detections or noise in tracking.






\section{Conclusion and Future Work}

In this paper, we propose a novel multi-object tracking method TNT based on a tracklet graph model, including tracklet vertex generation with epipolar geometry and connectivity edge measurement by a multi-scale TrackletNet. Our TNT outperforms other state-of-the-art methods on MOT16 and MOT17 benchmarks. We also show some qualitative results on different scenarios and applications using TNT. Robustness of TNT is further discussed with handling occlusions. 

However, fast camera motion is still a challenge in 2D tracking. 
In our future work, we are going to convert 2D tracking to 3D tracking with the help of visual odometry. Once the 3D location of the object in the world coordinate can be estimated, the trajectory of the object should be much smoother than the 2D case. 



{\small
\bibliographystyle{ieee}
\bibliography{egpaper_final}
}

\end{document}